\def\year{2021}\relax
\documentclass[letterpaper]{article} 
\usepackage{aaai21}  
\usepackage{times}  
\usepackage{helvet} 
\usepackage{courier}  
\usepackage[hyphens]{url}  
\usepackage{graphicx} 
\usepackage{natbib}  
\usepackage{caption} 
\usepackage{CJKutf8}
\usepackage{multirow}

\title{Predict and Use Latent Patterns for Short-Text Conversation}
\author{
    
    Hung-Ting Chen\footnote{indicates equal contribution}, 
    Yu-Chieh Chao\textsuperscript{\rm *}, 
    Ta-Hsuan Chao\textsuperscript{\rm *}, 
    Wei-Yun Ma
    
}

\affiliations{

    Academia Sinica \\
    \{timchen0618, vpj870331, alexchao2007, ma\}@iis.sinica.edu.tw

}

\begin{document}

\maketitle

\begin{abstract}
Many neural network models nowadays have achieved promising performances in Chit-chat settings. The majority of them rely on an encoder for understanding the post and a decoder for generating the response. Without given assigned semantics, the models lack the fine-grained control over
responses as the semantic mapping between posts and responses is hidden on the fly within the end-to-end manners. Some previous works utilize sampled latent words as a controllable semantic form to drive the generated response around the work, but few works attempt to use more complex semantic patterns to guide the generation. In this paper, we propose to use more detailed semantic forms, including latent responses and part-of-speech sequences sampled from the corresponding distributions, as the controllable semantics to guide the generation. Our results show that the richer semantics are not only able to provide informative and diverse responses, but also increase the overall performance of response quality, including fluency and coherence.
\end{abstract}

\section{Introduction}
The sequence-to-sequence neural network \cite{vinyals2015neural, shang2015neural} is one of the first successful neural models that generate text. However, the model are prone to producing generic and meaningless responses such as “I don’t know” or “So am I”.

To guide generation to produce more complex results, some works involve utilizing controllable semantics. Previous works \cite{ji2014information, yan2016learning, yan2016shall} try to generate human-like and relevant responses by editing retrieved responses. However, the downside is that these models are not trained end-to-end, which have to retrieve candidate responses for every new input post in order to produce a response. On the other hand, numerous prior studies \cite{xing2016topic, dziri2018augmenting, mou2016sequence, yao2017towards, gao2019generating} use keywords, either selected by neural networks or retrieval models, as semantic features to control the generation of the output text. As these methods mainly use simple semantic forms to assist generation, the semantic restriction is relatively limited and lack the global view on a whole generated sentence.

In this paper, we propose an end-to-end model that creates informative and diverse responses by providing more complex, sentence-wide semantics. We investigate two different latent patterns - latent sentences and latent part-of-speech (POS) sequences for guiding semantics. First, as many previous works use a word to assist dialogue generation, we use a complete sentence instead. Sentences provide pattern and word usage information to yield robust dialogue responses. Second, inspired by \citeauthor{shen2019controlling} (\citeyear{shen2019controlling}), who uses POS sequences to control the generation of Chinese poems, we serve POS sequences as hidden semantics to aid the Chit-chat task. We have shown that this method greatly helps improve the generation of dialogues, and that our generations indeed depend on the POS sequences we chose or generated.

Our model consists of two parts, the latent sequence predictor (including latent sentences and latent POS sequences), and the dialogue generator, which generates responses by the input post and the corresponding latent sequence we chose. After pre-training both parts, we combine the two networks into one end-to-end dialogue generation system. Finally, following \citeauthor{gao2019generating} (\citeyear{gao2019generating}), we fine-tune the whole network using a reinforcement learning (RL) algorithm. Our contributions can be summarized as follows:
\begin{itemize}
    \item We propose an end-to-end model with RL that predicts and utilizes complex latent patterns to guide the generation of dialogues with given input text. 
    \item We explore latent sentence and latent POS sequence as our semantic pattern to guide the generation, and find that the latter outperformed related state-of-the-art baselines.
    \item We implemented our dialogue generation models and all of our source codes and datasets are available at github. \footnote{https://github.com/timchen0618/LaPat}
\end{itemize}

\section{Related Work}
Many researchers have proposed methods to resolve the generic response problem. Maximum Mutual Information objective \cite{li2016diversity} and diverse beam search \cite{li2016simple} are used to address the issue. But these approaches only modify the generation at decoding steps. At training time, schemes such as reinforcement learning \cite{ranzato2015sequence, li2016deep, cuayahuitl2017simpleds} and adversarial training \cite{li2017adversarial, xu2017neural} are also applied.  One major drawback of these methods is that that they cannot explicitly control what to produce .

%
Regarding control of semantics in a response, two type of methods prevail in the dialogue generation domain, as presented in the following subsections.
\subsection{Rewriting and Editing a Sentence}
Retrieval methods \cite{ji2014information, yan2016learning, yan2016shall} have proved to produce more fluent outputs. Thus, many have proposed to rewrite the retrieved responses to balance between fluency and relevance to the input post. The candidates can serve as a good starting point, and then editing is performed to improve relevance. 

For instance, \citeauthor{song2016two} (\citeyear{song2016two}) combine retrieval and generative systems by referencing a candidate response through multi-source attention and post-reranking. \citeauthor{weston2018retrieve} (\citeyear{weston2018retrieve}) refine the retrieved response by treating it as an additional context. \citeauthor{wu2019response} (\citeyear{wu2019response}) treat the retrieved response as context and consider the lexical difference between the retrieved and the original post by augmenting the encoder-decoder model with an edit vector. Our work shares some similarities with these methods since we also reference a candidate response. However, we directly select a candidate response through neural models without doing retrieval first, eliminating the need for preprocessing of input data. Also, our model can be trained in an end-to-end manner and does inference as a whole. 

\subsection{Utilizing Latent Semantics Forms}
Others have sought to enhance the quality of responses by utilizing latent features hidden in the post-response pairs since they provide content or functional information. A wide variety of methods are built upon the help of a latent word. Two previous studies \cite{xing2016topic, dziri2018augmenting} select topic words with LDA (Latent Dirichlet Allocation) models and augment the encoder-decoder with topic-aware attention architecture. \citeauthor{mou2016sequence} (\citeyear{mou2016sequence}) and  \citeauthor{yao2017towards} (\citeyear{yao2017towards}) both select a latent word with the highest PMI (Pointwise Mutual Information) against the input post and use it to guide generation. \citeauthor{gao2019generating} (\citeyear{gao2019generating}) first select a latent word from the vocabulary and make the decoder attend to both the latent word and the input post.  \citeauthor{gao2019discrete} (\citeyear{gao2019discrete}) improve the CVAE structure by sampling a word from the vocabulary with a two-stage sampling scheme.

Researchers have also explored the possibility of using other latent semantic forms such as topic \cite{wang2017steering}, sentence function \cite{ke2018generating, bi2019fine}, frame semantics \cite{gupta2020controlling}, and lexical phrases \cite{wu2020controllable}. Previous works have mainly focused on using much simpler semantic forms, thus altering few words in the responses and might not be as effective. Instead, we attempt to guide the generation with more complex semantic forms, explicitly providing the model with patterns to follow suit. 

\section{Models}
The task we cope with in this paper is open-domain dialogue generation. An input post $\mathbf{p} \in \mathbf{P}$ is associated with multiple responses {$\mathbf{\{r\}}$}, where $\mathbf{r} \in \mathbf{R}$. Here $\mathbf{P}$ and $\mathbf{R}$ stands for the set of all the posts and the responses, respectively. We formulate the task as a sequence-to-sequence problem; given an input post $\mathbf{p} = \{p_1, ..., p_i, ..., p_n\}$ the model is expected to generate a response $\mathbf{r} = \{r_1, ..., r_t, ..., r_m\}$, where n and m represents the length of the post and response respectively. Latent semantic sequences $\mathbf{z} = \{z_1, ..., z_j, ..., z_l\}$ are introduced in the model. We denote latent sentences as $\mathbf{z_s} \in \mathbf{Z_s} = \mathbf{R}$ and latent part-of-speech tags as $\mathbf{z_p} \in \mathbf{Z_p}$, where $\mathbf{Z_p} = \{ pos( \mathbf{r}) | \forall \mathbf{r} \in \mathbf{R}\}$. 

\begin{figure}
    \centering
    \includegraphics[scale=0.215,height=32mm]{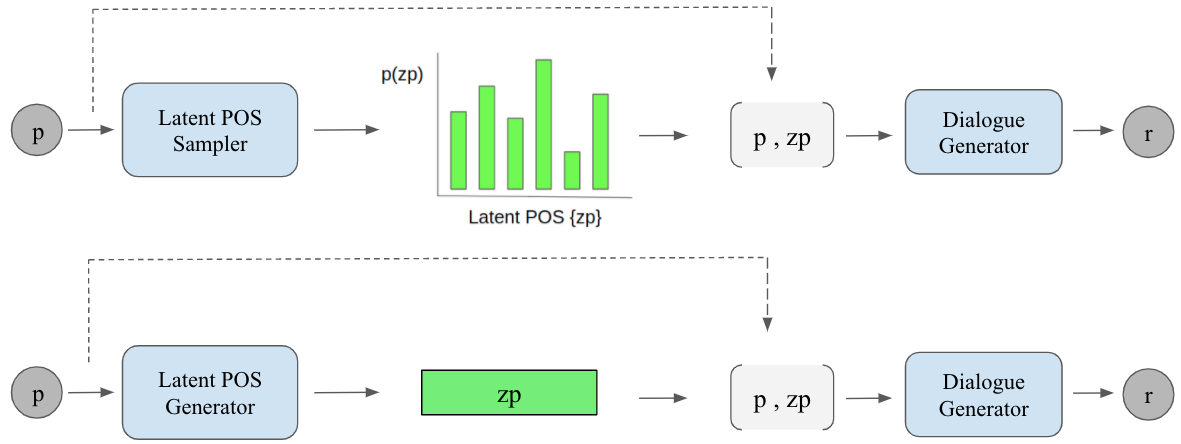}
    \caption{Overview of the proposed model. Latent sentence predictor share the architecture with latent POS sampler.}
    \label{fig:model_architecture_chart}
\end{figure}

Our model consists of two major components, as shown in figure~\ref{fig:model_architecture_chart}. The latent sequence predictor aims to predict the latent semantic sequence $\mathbf{z}$ conditioned on the post $\mathbf{p}$; the dialogue generator produces a response $\mathbf{r}$ according to $\mathbf{p}$ and $\mathbf{z}$.
Both models are first pretrained and then trained jointly in an end-to-end manner using reinforcement learning.

\subsection{Latent Sequence Predictor}

We tackle the selection of latent sentences and POS sequences as a classification problem. Due to the huge number of candidates, it is unlikely that we directly select one from the whole training dataset. Hence we construct a candidate set $\mathbf{\tilde{Z_i}} \subset \mathbf{Z_i} $ ($i \in \{s, p\}$) with $K_i$ latent sequences from the latent space $\mathbf{Z_i}$, as described in section~\ref{pre-data-prep}. Then for each post $\mathbf{p}$ we sample a $\mathbf{\tilde{z_i}}$ from the candidate set $\mathbf{\tilde{Z_i}}$ to guide the generation.
To produce more diverse POS sequences, we also propose a generation model for the Latent POS Predictor, which produces a POS sequence given the input post.

\subsubsection{Latent Sentence Predictor}

This network samples latent sentence $\mathbf{\tilde{z_s}}$ from candidate set $\mathbf{\tilde{Z_s}}$ by estimating $p(\mathbf{\tilde{z_s}}|\mathbf{p})$. We first encode the input post $\mathbf{p}$ with a bidirectional GRU (biGRU) to obtain an input representation $\mathbf{h^{p}}$, and retrieve hidden states of the last time step $t$ of the biGRU encoder to represent the meaning of the post:

\begin{equation} 
\mathbf{h_{t}^{p}} = [\overrightarrow{\mathbf{h_{t}^{p}}}, \overleftarrow{\mathbf{h_{t}^{p}}}] 
\end{equation}

and then compute the probability of each latent sentence:

\begin{equation}
 p(\mathbf{\tilde{z_s}}|\mathbf{p}) = softmax(\mathbf{MLP}(\mathbf{h_{t}^{p}})) 
\end{equation}
where parameters in the biGRU layer and Multilayer Perceptron (MLP) are trainable. $h_{t}^{p}$ denotes the hidden vector of last time step $t$ of the encoder.

\subsubsection{Latent POS Predictor}
Just like the latent sentence sampler, POS sequences can also be sampled from candidate set $\mathbf{\tilde{Z_p}}$. However, we also used a generation model for latent POS sequences to increase diversity. Next, we will elaborate on both the latent POS sampler and the latent POS generator.

\begin{itemize}
    \item Latent POS Sampler
    
    To sample a POS sequence, we simply use the Transformer \cite{vaswani2017attention} encoder to encode sentence:
    
    \begin{equation}
    \mathbf{h^p} = TransformerEncoder(\mathbf{p})
    \end{equation}  
    $$ \mathbf{h^{p}} = [\mathbf{h_{1}^p}, \mathbf{h_{2}^{p}}, ..., , \mathbf{h_{n}^{p}}] $$
    
    where $n$ is the length of the input post. Then we treat the last time step of the hidden vectors $h^{p}_{n}$ as our sentence representation. Finally, a MLP classifier is used to sample from the probability distribution of latent POS space:
    
    \begin{equation} p(\mathbf{\tilde{z_p}}|\mathbf{p}) = softmax(\mathbf{MLP}(\mathbf{h^{p}_{n}})) \end{equation}
    
    where parameters in the MLP block are trainable.
    
    \item Latent POS Generator
    
    Instead of sampling a POS sequence from all POS sequence candidates, we adopt the Transformer to generate a POS sequence $\mathbf{\tilde{z_{pg}}}$ based on the input post $\mathbf{p}$.
    
\end{itemize}

\subsection{Dialogue Generator}

This model aims to generate response $\textbf{r}$ from predicted latent semantics and input post $\textbf{p}$.

\subsubsection{Dialogue Generator with Latent Sentence}

With a sampled $\mathbf{\tilde{z_s}}$ from latent sentence predictor, we first encode both the input post $\mathbf{p}$ and the latent sentence $\mathbf{\tilde{z_s}}$ through two independent bidirectional GRU network for input representations $\mathbf{h_{p}^{g}}$ and $\mathbf{h_{\tilde{z_s}}^{g}}$ respectively. Both representations are then leveraged to decode the output sequence $\mathbf{y}$.



At each time step $t$, we add attention
implemented as in \cite{bahdanau2014neural}:

\begin{equation}
e_{i}^{t} = \mathbf{v^T}tanh(\mathbf{W_{h}} \mathbf{h^{g}_{i}} + \mathbf{W_{s}}\mathbf{s_{t}} + b_{attn})
\end{equation}
\begin{equation}
\mathbf{\alpha_{t}} = softmax(\mathbf{e_{t}})
\end{equation}
\begin{equation}
\mathbf{c_{t}} = \sum_{i}\alpha_{i}^{t} \mathbf{h^{g}_{i}} 
\end{equation}

where $\mathbf{v}$, $\mathbf{W_{h}}$, $\mathbf{W_{s}}$ and $b_{attn}$ are learnable parameters, and $\mathbf{c_{t}}$ is the context vector.
The decoder attends to encoding representation of both the input post and the latent sentence. The above context vector calculation works for both sources.

We adopt pointer generator network \cite{see2017get}. 
We calculate generation probability $\mathbf{p_{gen}}$ as:

\begin{equation} 
p_{gen} = \sigma(\mathbf{W_{gen}}[\mathbf{s_{t}}, \mathbf{c_{t}^{p}}, \mathbf{c_{t}^{\tilde{z_s}}}])
\end{equation}
\begin{equation}
l_{copy} = 1 - p_{gen}
\end{equation}

where $l_{copy}$ denotes copy probability (copying word from the latent sentence). $\mathbf{W_{gen}}$ is a learnable parameter, and $\sigma$ is the sigmoid function. $\mathbf{c_{t}^{p}}$ represents the input post context vector and $c_{t}^{\tilde{z_s}}$ represents the latent sentence context vector.

Finally, the probability distribution over the extended vocabulary (pre-set vocabulary \& OOV words) is:

\begin{equation} 
P(w) = p_{gen}P_{vocab}(w) + (1-p_{gen})\sum_{i:w_{i}=w}a_{i,t}^{\tilde{z_s}} 
\end{equation}

where $a_{i,t}^{\tilde{z_s}}$ denotes latent sentence attention distribution. Note that if $w$ is an out-of-vocabulary (OOV) word, then $P_{vocab}(w)$, which represents the probability distribution over pre-set vocabulary, is zero.

\subsubsection{Dialogue Generator with POS sequence}

Inspired by \cite{shen2019controlling}, we concatenate the predicted POS sequence $ \mathbf{\tilde{z_p}}$ (or in the generated POS sequence case, $ \mathbf{\tilde{z_{pg}}}$) right behind the post at the input layer. Next, we forward the concatenated form into Transformer architecture as in \cite{vaswani2017attention} to generate the corresponding response:

\begin{equation}
\mathbf{\hat{r}} = Transformer([\mathbf{p}, \mathbf{\tilde{z_p}}])
\end{equation} 

The technique is capable of producing sentences matching the input POS sequences, hence providing enough guidance for generating responses with valid structure.

\subsection{Pretraining and Data Preparation} \label{pre-data-prep}
We first pretrain the latent sequence predictor and the dialogue generator separately. For the latent sentence predictor and latent POS sampler, they are trained to solve a sentence classification problem. The creation of data for pretraining is explained in details below: 

\subsubsection{Candidate Set of Latent Sentences}
We attempt to construct a candidate set of size $K_s$ representative enough of the entire response set $\mathbf{R}$. First, we use kmeans-clustering to aggregate responses into $C$ clusters, and select $K_s/C$ responses from each cluster. We then assign an index, i.e., label for classification, to each of the latent responses in $\mathbf{\tilde{Z_s}}$. Finally, we assign the label of $\mathbf{z_s^*}$ to each $(\mathbf{p},\mathbf{r})$, where $\mathbf{z_s^*}$ is the most similar response in $\mathbf{\tilde{Z_s}}$ with the target response $\mathbf{r}$. The similarity between responses is defined as the Euclidean distance of the BERT\footnote{https://github.com/hanxiao/bert-as-service}  sentence encoding of the two.
\subsubsection{Candidate Set of Latent POS Sequences}
First, we derive the part-of-speech tagging of each response in $\mathbf{R}$ using Jiagu\footnote{https://github.com/ownthink/Jiagu}. Likewise, the candidate POS sequences must be representative enough of $\mathbf{Z_p}$. Thus we pick the $K_p$ most common POS sequences in $\mathbf{Z_p}$ as our candidate set $\mathbf{\tilde{Z_p}}$. 

Then, we create the training data for the latent POS sampler. We first assign classification label to each of the $\mathbf{z_p} \in \mathbf{\tilde{Z_p}}$. For each input post-response pair $(\mathbf{p},\mathbf{r})$, we find the most similar $\mathbf{z_p^*}$ to the POS sequence of r and label the pair with the corresponding index of $\mathbf{z_p^*}$. That is, we wish to predict the class label of $\mathbf{z_p^*}$ given the input post $\mathbf{p}$. The similarity between POS sequences is calculated by alignment.\footnote{https://biopython.org/DIST/docs/api/Bio.pairwise2-module.html}

\medskip
The generation of POS sequences and responses are both considered sequence-to-sequence generation problems. The latent semantic generator reads in the input post and outputs a POS sequence. For the dialogue generator, we simply concatenate the POS sequence $\mathbf{z_p}$ of response $\mathbf{r}$ after the post $\mathbf{p}$ and input the newly created sequence [$\mathbf{p}$;$\mathbf{z_p}$] into the Seq2Seq model.

\subsection{Joint Training With Reinforcement Learning}
To acquire better latent semantic sequences for the generation model, we fine-tune both models end-to-end with a reinforcement-learning objective. We treat the latent sequence predictor as an agent and the predictions of latent semantics as actions. We intend to choose the policy which enables the model to gain maximum total rewards. Note that during joint training, we apply the reinforcement learning algorithm only on the latent sequence predictor, whereas the dialogue generator is directly optimized through standard cross-entropy loss. 
We apply the REINFORCE \cite{williams1992simple} algorithm, a Monte-Carlo based policy gradient method; the formula of updating parameters of the latent sequence predictor (denoted as $\theta$ ) is as follows. $ Q(\mathbf{\hat{z}}) $ stands for estimated return if we sample a latent $ \mathbf{\hat{z}} $ ($ \mathbf{\hat{z}} $ can be either a latent sentence or a latent POS sequence), and we only calculate the return after the full response is generated. 
\begin{itemize}
    \item For selecting a sequence (either a sentence or a POS sequence)
\end{itemize}
\begin{equation} 
\theta \leftarrow \theta +  \nabla_{\theta}{Q(\mathbf{\hat{z}})\log{p_{\theta}(\mathbf{\hat{z}}|\mathbf{p})}} 
\end{equation}

\begin{itemize}
    \item For generating a POS sequence
\end{itemize}
\begin{equation} 
\theta \leftarrow \theta +  \sum_{j=0}^{l}{\nabla_{\theta}{Q(\mathbf{\hat{z^p}})\log{p_{\theta}(\mathbf{\hat{z^p_{j}}}|\mathbf{p})}}} 
\end{equation}
We model the estimated return as a reward function measuring the similarity between the predicted and target response. We update the latent sequence predictor only according to the maximum reward acquired from the bag of responses. We found empirically that it is challenging for the model to match all the responses during joint training since it has to explore the large latent space to select appropriate latent semantic sequences for all the responses. The reward function $Q(\mathbf{\hat{z}})$ is designed as follows; here $\mathbf{\{r\}}$ denote the set of multiple target responses associated with the post $\mathbf{p}$ 

\begin{equation}
Q(\mathbf{\hat{z}}) = R(\bf{\hat{r}}, \bf{\{r\}}) = \max\limits_{\bf{r} \in \{\bf{r}\}}R(\bf{\hat{r}}, \bf{r})
\end{equation}
, where 
\begin{equation} 
\mathbf{R}(\mathbf{\hat{r}}, \mathbf{r}) = F1(\mathbf{\hat{r}}, \mathbf{r}) \end{equation}
Here, a variation of F1 score calculates the overlap between the generated response and the target. Results have shown the joint training scheme enhances overall response quality.


\section{Experiments}
\begin{table*}
\begin{center}
\begin{tabular}{ |c|c|c|c|c| } 
 \hline
 Models & BLEU-1 & BLEU-2 & BLEU-3 & BLEU-4 \\
 \hline
 Seq2Seq & 25.57 & 7.39 & 1.71 & 0.64 \\
 HGFU \cite{yao2017towards} & 27.31 & 9.59 & \bf{4.08} & \bf{2.14} \\
 ResponseEdit \cite{wu2019response} & 24.45 & 7.32 & 2.38 & 0.96 \\
 CVAE \cite{zhao2017learning} & 20.13 & 6.82 & 1.36 & 0.52 \\ 
 GMDR \cite{gao2019generating} & 21.26 & 8.27 & 3.93 & 2.08 \\
 \hline
 \hline
 Ours-Two Stage (latent sentence) & 17.19 & 4.72 & 0.42 & 0.10 \\ 
 Ours-Two Stage (sample POS) & \bf{22.28} & \bf{5.31} & 1.11 & 0.00 \\ 
 Ours-Two Stage (generate POS) & 22.84 & 6.08 & 1.07 & 0.00 \\ 
 \hline
 Ours-RL (latent sentence) & 19.21 & 6.83 & 1.61 & 0.23 \\ 
 Ours-RL (sample POS) & \bf{36.42} & \bf{12.35} & 3.55 & 0.99 \\ 
 Ours.  RL  (generate POS) & 30.98 & 10.21 & 3.25 & 1.29 \\ 
 \hline
\end{tabular}

\caption{Automatic evaluation results on Weibo.}
\label{table:evaluation_weibo}
\end{center}
\end{table*}

\subsection{Dataset}
Our method is evaluated on the Weibo Bechmark Dataset, an open-domain Chinese dialogue dataset with over 400 million training pairs. The testing set includes 3,200 data pairs. Here, we use the modified version released by \cite{gao2019generating}. \footnote{https://ai.tencent.com/ailab/nlp/dialogue.html}

\subsection{Evaluation}
We report the results of the BLEU score \cite{papineni2002bleu}, a widely used metric measuring the similarity between the predicted response and the ground truth. We provide the values of BLEU 1-4 for all the experiments. 
 
However, since the automatic evaluation metrics cannot effectively reflect the goodness of dialogue generation, we also did human evaluation on  100 randomly selected sentences from our and other comparing models. Three native speakers are asked to label the responses according to fluency, relevance and informativeness (generic responses would get low score in this category). Each response is given a score of 1 to 5 according to these quality factors.


\subsection{Baseline Methods}
We compare our method with popular models as baseline methods. The vanilla Seq2Seq model \cite{vinyals2015neural}; HGFU \cite{yao2017towards}, a model which modifies GRU by incorporating a cue word in the generation process; ResponseEdit \cite{wu2019response}, a retrieve-and-edit model utilizing an edit vector; CVAE \cite{zhao2017learning}, and GMDR \cite{gao2019generating}, a latent word inference network combined with a joint-attention generation network.




\subsection{Implementation Details}
\subsubsection{Data Preparation}

\begin{itemize}
    \item Latent Sentence
\end{itemize}

We set $C$ = 1000 and $K_s$ = 50000 for the construction of the candidate set $\mathbf{\tilde{Zs}}$. The pretraining data for the latent sentence sampler thus includes input post $\mathbf{P}$ as input and distribution of $\mathbf{Z_s^*}$  as output.

\begin{itemize}
    \item Latent POS Sequence  
\end{itemize}



We expect the candidate set $\mathbf{\tilde{Zp}}$ to be representative enough of $\mathbf{Zp}$.  In our experiments, we choose $K_p$ = 500, 1000, and 10000 to test how different size of candidate set affect the performance of our model.

\subsubsection{Training Details}
\begin{itemize}
    \item Models Using Latent Sentences
\end{itemize}

%
In latent sentence models, we utilize GRU as our fundamental models. For the latent sentence predictor, we adopt 1-layer biGRU as our sentence encoder, and a classifier with $3$ fully-connected layers. For the Dialogue Generator, we use 1-layer biGRU for the encoders of both input posts and latent sentences. A decoder of $1$ layer uni-directional GRU is used. The dimensions are 1024 for the encoders and 512 for the decoder and classifier.

All models are trained with Adam optimizer. The learning rate for the latent sentence predictor and the dialogue generator are set to $0.002$ and $0.0002$, respectively. The learning rate decay is $0.5$ each epoch. During testing, we apply beam-search with beam size 4.
\begin{itemize}
    \item Models Using Latent POS Sequences
\end{itemize}
For experiment utilizing latent POS sequences, the dialogue generator is a standard $6$-layer Transformer encoder-decoder model. We follow the model architecture and the parameter of the base model in \citeauthor{vaswani2017attention} (\citeyear{vaswani2017attention}).
The latent POS sampler model is a 6-layer transformer encoder with a  classifier the same as the Latent Sentecnce Predictor on top of the last time step of hidden states. The latent POS generator shares the model architecture with the dialogue generator. 

All Transformer models are trained with the Adam optimizer. For the pretrained models, we adopt noam learning rate decay as proposed by \citeauthor{vaswani2017attention} (\citeyear{vaswani2017attention}); the warmup step is set to $8000$. The initial learning rate is set to $\mathbf{10^{-5}}$ for joint fine-tuning, and is decayed every epoch. During testing, the beam size is set to 3.





\section{Results}

\subsection{Automatic Evaluation}

The evaluation results of our model are shown at Table~\ref{table:evaluation_weibo}. As we can see, our Ours-RL(sample POS) outperforms other models in BLEU 1 and 2. The reason is that the sentence patterns we provided narrow down the word selection at each time step. For example, if the corresponding POS tag for the token at this time step is $v$, then the model will generate a verb with higher probability. Thus, our model can choose more accurate and relevant words, and thus achieves outstanding performance in BLEU 1 and 2. However, since our model forces the pattern of our generation (which is normally different from the original response), it does not perform as well in longer word matching (BLEU 3 and 4). 

We also report the performance of combining the pretrained models as Ours-Two Stage, i.e., directly using the latent sequence predicted by pretrained latent sequence predictor during generation. The BLEU scores improved significantly after fine-tuning with our reinforcement learning algorithm since both modules are optimized in an end-to-end manner, enabling more appropriate prediction of latent sequences. The BLEU score of Ours-RL(sample POS) improved \textbf{63\%} and \textbf{132\%} and achieved \textbf{36.42} and \textbf{12.35} in BLEU-1 and BLEU-2 scores.


However, our latent sentence model does not yield good BLEU scores because the generation is greatly influenced by the chosen latent sentence, as shown in Table~\ref{table:ngram-overlap}. Thus, if the latent sentence we have chosen is very different from the original response, the BLEU scores would drop. Also, the search space for the latent sentence sampler is way smaller than the ResponseEdit model, which uses pre-define index to search for similar posts in all posts at test time. Possible future work could be increasing the latent space and add some pre-defined constraints during decoding.
\begin{table}
\begin{center}
\begin{tabular}{ |c|c|c|c| } 
 \hline
 Models & Fluency & Relevance & Informativeness  \\
 \hline
 HGFU & 3.69 & 2.65 & 2.60\\
 \hline
 GMDR & \bf{4.51} & 3.03 & 2.73\\
 \hline
 Ours(select POS) & 3.98 & \bf{3.36} & \bf{3.34}\\
 \hline
\end{tabular}
\caption{Result of Human Evaluation.}
\label{table:human_eval}
\end{center}
\end{table}
\subsection{Human Evaluation}
The result of human evaluation is shown in Table~\ref{table:human_eval}.  The model that we use for comparison is latent POS sampler and $K_p$ = 500.
Our model obtains the highest scores in both relevance and informativeness. The reason is that our model incoporates latent patterns, which provide guidance signal about how to organize content and when to include certain information. With a reasonable pattern to follow, the generated responses are thus more relevant and informative.  However, since the outputs of the generated sentences follow the POS patterns, they might sacrifice fluency slightly. The fluency score is still better than HGFU, though.

\subsection{Case Study and Analysis}
\begin{CJK*}{UTF8}{gbsn}
 Table~\ref{table:case_study} shows one of the dialogue predicted by our models and other models. The model GMDR \cite{gao2019generating} predicted a latent word \textit{darling}, and thus generated a simple sentence \textit{All of them are my darlings}. However, our models produced more complex results, both in word usage and sentence patterns. Our latent sentence model chose \textit{The sky of the liberated area is a sunny day.} as the latent sentence. (The sentences in brackets are selected latent responses.) Thus, it affected the generation \textit{Oh my god} (because in Chinese, \textit{day 天} is equal to \textit{god 天}). As for the latent-POS-predicting models, the selected POS sequence provided a reasonable output sentence pattern for the model to follow. Thus, we obtain an informative and relevant response \textit{They are so cute, I'll be blessed, too. I miss my darling so much}. 
\end{CJK*}
 
\begin{table}
\begin{center}
\begin{CJK*}{UTF8}{gbsn}
\small
\begin{tabular}{ |c|c|c|c| } 
 \hline
 \multirow{2}{5.5em}{Input} & 这些 都 是 我 的 宝贝 们 \\ 
 & All of these are my darlings. \\
 \hline
 \multirow{2}{5.5em}{Seq2Seq} & 我 也 想 说 ， 我 也 想 吃 了 。 \\ 
 & I also want to say; I want to eat, too. \\
 \hline
 \multirow{2}{5.5em}{HGFU} & 都 是 我 喜欢 的 ， 好 可爱 \\ 
 & All of them are my favorite, they're so cute. \\
 \hline
 \multirow{2}{5.5em}{GMDR} & 都 是 我 的 宝贝 们 。  \\ 
 & All of them are my darlings. \\
 \hline
 \multirow{2}{5.5em}{CVAE} & 还有 那个 可以 吗 ？  \\ 
 & Is that available? \\
 \hline
 \multirow{4}{5.5em}{Ours(Latent Sentence)} &  我的 天 我的 天 ， 喜欢 。  \\ 
 & [解放区 的 天 是 晴朗 的 天 ] \\
 & Oh my god oh my god, like it. \\ 
 & [The sky of the liberated area is a sunny day. ] \\
 \hline
 \multirow{2}{5.5em}{Ours(Generate POS)} & 好 可爱 啊 ， 好 可爱 ， 好 喜欢 。   \\ 
 & So cute, so cute, and I like it so much.  \\
 
 \hline
 \multirow{3}{5.5em}{Ours(Sample POS)} & 好 可爱 啊 ， 我 也 要 幸福 的 。 好 想 我 的 宝宝 啊 。  \\ 
 & They are so cute, I'll be blessed, too. I miss\\
 & my darling so much.  \\
 \hline
\end{tabular}

\caption{Evaluation results of different sizes of latent space}
\label{table:case_study}
\end{CJK*}
\end{center}
\end{table}

\subsection{Analysis On the Selection of Latent Space}
 Table~\ref{table:diff_latent_space} shows the results of how different latent space size $K_p$ affect the performance of our model. We can see that the best results occur at $K_p$ = 500. If the latent spaces is too large, then it is difficult for the model to search through the whole space for appropriate latent sequences.

\begin{table}
\begin{center}
\begin{tabular}{ |c|c|c|c|c| } 
 \hline
 $K_p$ & BLEU-1 & BLEU-2 & BLEU-3 & BLEU-4 \\
 \hline
 \hline
 500 & 36.42 & 12.35 & 3.55 & 1.29 \\ 
 1000 & 28.60 & 9.12 & 3.09 & 1.07 \\ 
 10000 & 22.17 & 7.51 & 2.70 & 1.03 \\ 
 \hline
\end{tabular}

\caption{Evaluation results of different sizes of latent space}
\label{table:diff_latent_space}
\end{center}
\end{table}

\subsection{Faithfulness to Given Semantic Sequences} \label{faithfulness}
To prove that the selected latent semantic sequence is actually helpful for generation, we inspect if the generated response follows the provided word or POS sequence. 

For model utilizing latent sentences, we inquire whether the model mimics the word selection (providing content) and the word ordering (providing word usage, sentence pattern, and content ordering) of the given latent sentence. As for word selection, we calculate the percentage of overlapping words with the latent sentences in generated responses. For word order, we calculate the percentage of overlapping n-grams (n$\geq$2). The result is presented in table \ref{table:ngram-overlap}.

\begin{figure}
    \centering
    \includegraphics[scale=0.25,height=35mm]{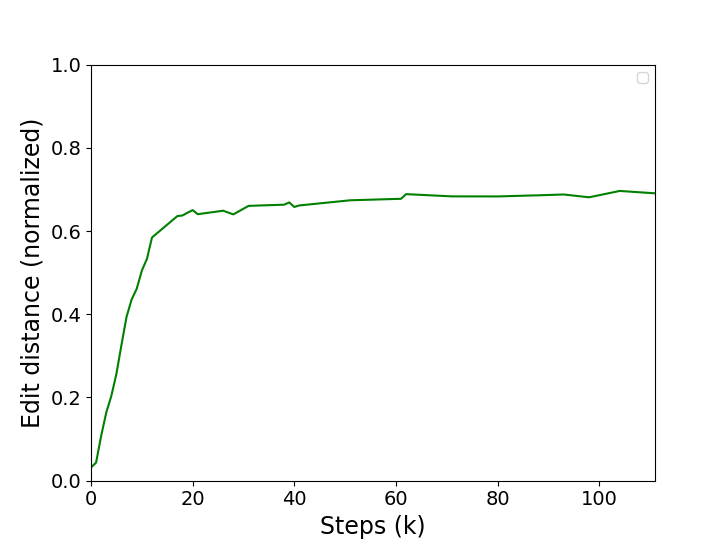}
    \caption{Edit distance during fine-tuning process of RL.}
    \label{fig:my_label}
\end{figure}

\begin{table} 
\begin{center}

\begin{tabular}{ |c|c|c|c|c| } 
 \hline
 & unigram & bigram & trigram & 4-gram  \\
 \hline
  overlap (\%) & 71.11 & 44.86 & 32.84 & 26.76   \\
 \hline
\end{tabular}

\caption{N-gram overlap - latent sentences and responses}
\label{table:ngram-overlap}
\end{center}
\end{table}

As can be seen from the result, the model copies over two-third of words and about one-forth of the words from the latent sentence. It clearly illustrates that the model indeed referencs to the latent sentence for better response generation. However, the model might also suffers from copying bad latent sentences. Thus we believe the overall response quality will improve if we could ensure better latent sentence selection, which may be a potential future study. 

For models using latent POS sequences, the output responses mostly follow the given POS sequence during pretraining but learns to generate more freely when fine-tuning. We believe our model not only follows the guidance of the underlying patterns of latent POS, but also maintains a certain extent of freedom concerning the generation of word sequences. To support the claim, we measure the faithfulness of generation to the provided POS sequence using the edit distance between the POS of output responses and the given POS sequence. Smaller edit distance indicates greater faithfulness to the latent POS. The edit distance is normalized by the length of the selected POS sequence to simulate the mismatch percentage.

Figure~\ref{fig:my_label} shows the result during fine-tuning for our best model. The model first strictly follows the selected latent POS sequence but later begins to generate responses more freely. We thus prove statistically that our model indeed strikes a balance between generating word segments on its own and referencing to the latent POS sequence.    
The automatic evaluation results also proves our theory, since fine-tuned models greatly outperform the two-stage models.

\section{Conclusion}
We propose an end-to-end response generation network aided by complex latent semantics. Our model uses latent sentences and latent POS sequences respectively to provide wording and sentence pattern information. Also, our combined end-to-end model can be optimized by a reinforcement learning algorithm, and is capable of striking a balance between totally and partially following the latent semantics during generation. Our results show that our model increases the quality of responses by being more informative and relevant to the input post compared with existing baselines.

\section{Future Work}

Our next step would be using a mixed latent sequence to help guide the generation. For example, we can integrate both POS tags and words into a sequence to help our model generate more informative and human-like results. Also, we could try other semantic forms to assist our generation, such as Semantic Role Labeling (SRL).

\nocite{*} 

\end{document}